\documentclass[10pt,twocolumn,letterpaper]{article}

\usepackage[pagenumbers]{cvpr} %

\usepackage[dvipsnames]{xcolor}

\definecolor{cvprblue}{rgb}{0.21,0.49,0.74}
\usepackage[pagebackref,breaklinks,colorlinks,citecolor=cvprblue]{hyperref}

\usepackage{amsmath}
\usepackage{amssymb}
\usepackage{mathtools}
\usepackage{amsthm}
\usepackage{amsmath}

\usepackage{microtype}

\usepackage{inconsolata}

\usepackage{graphicx}
\usepackage{booktabs} %
\usepackage{tabularx}
\usepackage{lipsum}  
\usepackage{listings}
\usepackage{caption}
\usepackage{subcaption}
\usepackage{CJKutf8}
\usepackage{xspace}
\usepackage{listings}
\usepackage{xcolor}
\usepackage{float}

\definecolor{codegreen}{rgb}{0,0.6,0}
\definecolor{codegray}{rgb}{0.5,0.5,0.5}
\definecolor{codepurple}{rgb}{0.58,0,0.82}
\definecolor{backcolour}{rgb}{0.95,0.95,0.92}

\lstdefinestyle{mystyle}{
    backgroundcolor=\color{backcolour},   
    commentstyle=\color{codegreen},
    keywordstyle=\color{magenta},
    numberstyle=\tiny\color{codegray},
    stringstyle=\color{codepurple},
    basicstyle=\ttfamily\footnotesize,
    breakatwhitespace=false,         
    breaklines=true,                 
    captionpos=b,                    
    keepspaces=true,                 
    numbers=left,                    
    numbersep=5pt,                  
    showspaces=false,                
    showstringspaces=false,
    showtabs=false,                  
    tabsize=2
}

\newfloat{lstfloat}{htbp}{lop}
\floatname{lstfloat}{Algorithm}

\lstdefinestyle{myverbatim}{
    basicstyle=\ttfamily\footnotesize,
    backgroundcolor=\color{white},
    breaklines=true,
    breakatwhitespace=true
}

\usepackage{multirow}
\usepackage[framemethod=TikZ]{mdframed}
\usepackage{tabu}
\usepackage{tcolorbox}
\usepackage{soul}
\usepackage{makecell}

\definecolor{cb-0}{RGB}{216, 27, 96}
\definecolor{cb-1}{RGB}{30,136,229}
\definecolor{cb-2}{RGB}{255,193,7}
\definecolor{cb-3}{RGB}{0, 77, 64}
\definecolor{cb-4}{RGB}{150,220,174}

\newcommand{\ourmethodshortname}{\textit{SESAME}}
\newcommand{\chainedmethod}{Cascading}
\newcommand{\datasetname}{FP-RefCOCO}
\newcommand{\datasetnameplain}{FP-RefCOCO}
\newcommand{\datasetnameplus}{FP-RefCOCO+}
\newcommand{\datasetnameg}{FP-RefCOCOg}

\def\paperID{4164} %
\def\confName{CVPR}
\def\confYear{2024}

\title{See, Say, and Segment: Teaching LMMs to Overcome False Premises}

\author{Tsung-Han Wu\footnotemark[1] \quad Giscard Biamby\footnotemark[1] \quad David Chan \quad Lisa Dunlap \\
Ritwik Gupta \quad Xudong Wang \quad Joseph E. Gonzalez \quad Trevor Darrell\\
University of California, Berkeley \\
}

\begin{document}
\maketitle
\footnotetext[1]{*Equal contribution.}
\def\tabfpref#1{
\begin{table*}[#1]
    \centering
    \tabcolsep=0.18cm
    {
        \begin{tabular}{ l | c c c | c c c | c c c }
            \toprule
            
            \multirow{3}*{Method} & \multicolumn{3}{c|}{\datasetname{}} & \multicolumn{3}{c|}{\datasetnameplus{}}  & \multicolumn{3}{c}{\datasetnameg{}} \\ 
            
            \specialrule{0em}{0pt}{1pt}
            \cline{2-10}
            \specialrule{0em}{0pt}{1pt}
            
            ~ & See & Say & Segment & See & Say & Segment & See & Say & Segment \\

            \specialrule{0em}{0pt}{1pt}
            \hline
            \specialrule{0em}{0pt}{1pt}

            LISA & 51.36 & 0.00 & 44.00 & 51.32 & 0.00 & 39.62 & 51.25 & 0.00 & 39.64 \\
            \chainedmethod{} (Ours) & 75.59 & 0.35 & 55.18 & 75.03 & 0.42 & 48.64 & 76.07 & 0.55 & 49.98\\
            \ourmethodshortname{} (Ours) & \textbf{79.84} & \textbf{0.63} & \textbf{57.93} & \textbf{80.00} & \textbf{0.61} & \textbf{50.81} & \textbf{81.78} & \textbf{0.67} & \textbf{53.79}\\
            
            \bottomrule            
        \end{tabular}
    }
    \caption{SESAME (ours) and existing methods on various referring segmentation tasks.  The See scores measure the binary classification accuracy. Say is measured via CLAIR score, which ranks the similarity of the suggested false premise correction against positive referring expressions for the same referent. The Segment scores are (cIoU).}
    \label{table:refer_seg}   
\end{table*}
}

\def\tabAbl#1{
\begin{table*}[#1]
    \centering
    \tabcolsep=0.18cm
    {
        \begin{tabular}{ l | c c | c c c c }
            \toprule
            
            \multirow{3}*{Method} & \multicolumn{2}{c|}{See} & \multicolumn{4}{c}{Segment (cIoU)}   \\ 
            
            \specialrule{0em}{0pt}{1pt}
            \cline{2-7}
            \specialrule{0em}{0pt}{1pt}
            
            ~ & Recall (FP) & Recall (TP) & 0\% FP & 25\% FP & 50\% FP & 75\% FP  \\

            \specialrule{0em}{0pt}{1pt}
            \hline
            \specialrule{0em}{0pt}{1pt}

            LISA & ~~0.00 & \textbf{100.00} & \textbf{67.99} & 52.36 & 39.64 & 34.15 \\
            \chainedmethod{} (Ours) & 58.76 & ~~94.64 & 65.77 & 58.53 & 49.98 & 45.82\\
            \ourmethodshortname{} (Ours) & \textbf{67.89} & ~~96.64 & 66.02 & \textbf{60.64} & \textbf{53.79} & \textbf{49.79} \\
            
            \bottomrule            
        \end{tabular}
    }
    \caption{Ablation on the amount of false premise data used at test time, for See and Segment scores on the \datasetnameg{} dataset. \ourmethodshortname{} (ours) demonstrates superior segmentation performance even the false premise sampling rate is as high as 75\%.}
    \label{table:abl}   
\vspace{-0.2cm}
\end{table*}
}

\def\tabNatref#1{
\begin{table}[#1]
    \centering
    \tabcolsep=0.18cm
    {
        \begin{tabular}{ l | c  c  c }
            \toprule
            
            Method & refCOCO & refCOCO+ & refCOCOg \\ 
            \midrule

            MCN~\citep{luo2020multi} & 62.4 & 50.6 & 49.2  \\

            VLT~\citep{ding2021vision} & 67.5 & 56.3 & 55.0 \\

            CRIS~\citep{wang2022cris} & 70.5 & 62.3 & 59.9  \\

            LAVT~\citep{yang2022lavt} & 72.7 & 62.1 & 61.2 \\
            
            ReLA~\citep{liu2023gres} & 73.8 & \textbf{66.0} & 65.0 \\
            
            X-Decoder~\citep{zou2023generalized} & - & - & 64.6   \\

            SEEM~\citep{zou2023segment} & - & - & 65.7   \\
            
            LISA-7B \cite{lai2023lisa} & \textbf{74.9} & 65.1 & \textbf{67.9}  \\
            \midrule
            \ourmethodshortname{} (Ours) & 74.7 & 64.9 & 66.1 \\
            
            \bottomrule            
        \end{tabular}
    }
    \caption{Even though our method is trained to do both see, say, and segment simultaneously, our model is still on par with prior methods on natural setting.}
    \label{tab:segmentations}   
\vspace{-1em}
\end{table}
}

\def\tabSupp#1{
\begin{table}[#1]
    \centering
    \setlength\tabcolsep{0.18cm}\resizebox{1.0\linewidth}{!}
    {
        \begin{tabular}{ l | c c | c}
            \toprule
            
            \multirow{3}*{Method} & \multicolumn{2}{c|}{See } & \multicolumn{1}{c}{Segment}   \\ 
            
            \specialrule{0em}{0pt}{1pt}
            \cline{2-4}
            \specialrule{0em}{0pt}{1pt}
            
            ~ & FP Recall & TP Recall & cIoU \\

            \specialrule{0em}{0pt}{1pt}
            \hline
            \specialrule{0em}{0pt}{1pt}
            LISA & 0.5 & \textbf{100.0} & 42.40 \\
            \chainedmethod{} (Ours) & 7.0 & 99.5 &  44.96 \\
            \ourmethodshortname{} (Ours) & \textbf{86.5} & 90.0 &  \textbf{51.43}  \\
            \bottomrule            
        \end{tabular}
    }
    \caption{In addition to referring segmentation tasks detailed in \cref{table:refer_seg} of our main paper, \ourmethodshortname{} (ours) also exhibits significant gain in reasoning segmentation tasks, hugely surpasses both our chained model methods and the LISA baseline by a large margin.}
    \label{table:reasoning_seg}   
\vspace{-0.2cm}
\end{table}
}

\begin{abstract}

Current open-source Large Multimodal Models (LMMs) excel at tasks such as open-vocabulary language grounding and segmentation but can suffer under false premises when queries imply the existence of something that is not actually present in the image. 
We observe that existing methods that fine-tune an LMM to segment images significantly degrade their ability to reliably determine (``see'') if an object is present and to interact naturally with humans (``say''), a form of catastrophic forgetting. In this work, we propose a cascading and joint training approach for LMMs to solve this task, avoiding catastrophic forgetting of previous skills. Our resulting model can ``see'' by detecting whether objects are present in an image, ``say'' by telling the user if they are not, proposing alternative queries or correcting semantic errors in the query, and finally ``segment'' by outputting the mask of the desired objects if they exist. Additionally, we introduce a novel False Premise Correction benchmark dataset, an extension of existing RefCOCO(+/g) referring segmentation datasets (which we call \datasetname(+/g)). The results show that our method not only detects false premises up to 55\% better than existing approaches, but under false premise conditions produces relative cIOU improvements of more than 31\% over baselines, and produces natural language feedback judged helpful up to 67\% of the time.
\end{abstract}    

\section{Introduction}

Perception systems engaging with real-world environments often need to understand and respond to complex queries such as ``find the keys with the purple heart on them'' or ``bring me the remote for the television.'' Solving such complex visual tasks can require active reasoning, world knowledge, and an implicit understanding of the scene which are often unavailable to simple visual perception systems \cite{lai2023lisa}. An extension of referring segmentation \cite{hu2016segmentation}, ``reasoning segmentation,'' requires that models are capable not only of understanding the query but reasoning on the query as well. 

\begin{figure}[t]
    \centering\vspace{4ex}
    \includegraphics[width=\linewidth]{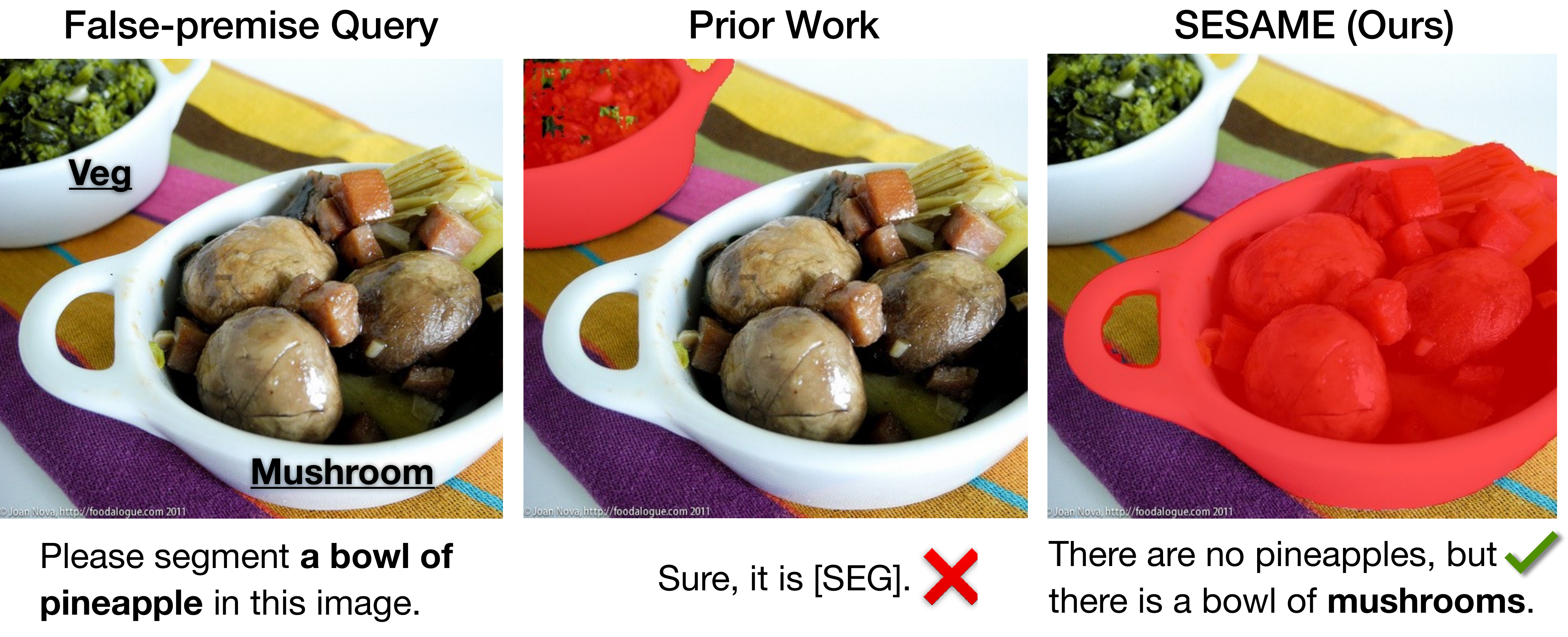}
    \caption{False premise failures with LMMs: contemporary open-source LMMs combined with segmentation decoders are able to generate referring segments effectively but have difficulty on segmentation questions which ask the model to refer to something that is not present in the image. SESAME, our See-Say-Segment LMM, uses model chaining and joint training to overcome this problem. }
    \vspace{-1em}
    \label{fig:teaser}
\end{figure}

\begin{figure*}[t]
    \vspace{2ex}
    \centerline{\includegraphics[width=1.00\linewidth]{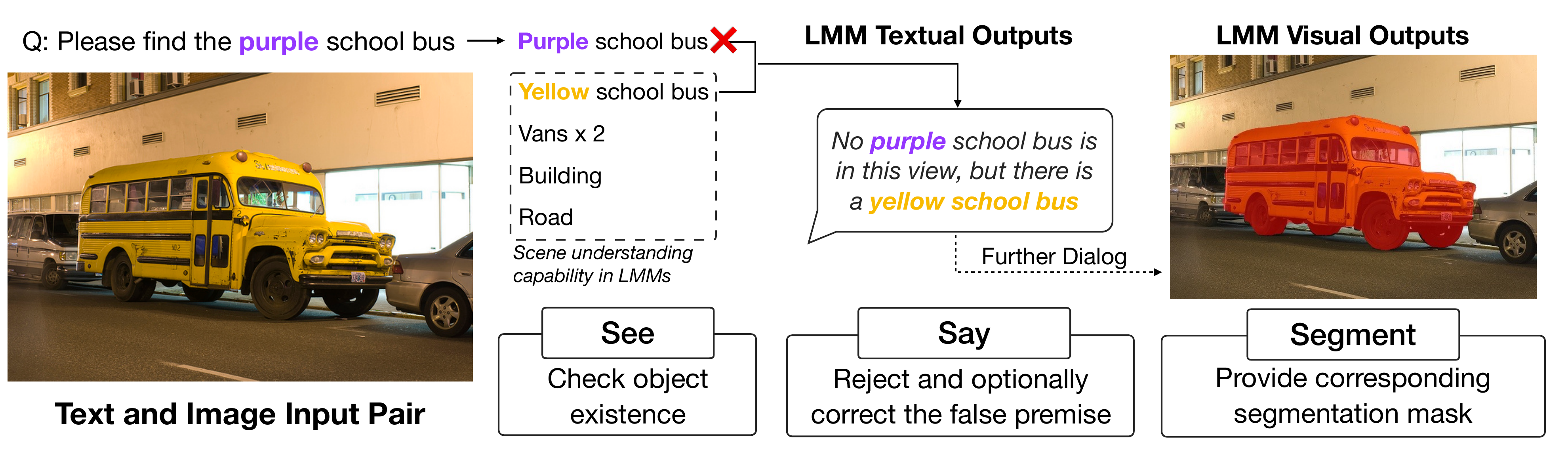}}
    \caption{\ourmethodshortname{} is an LMM that can ``see'' whether objects are detected in an image and ``say'' by telling the user if they are there or not. When appropriate, alternative queries can be offered or semantic errors corrected in the query.  SEASAME can then ``segment'' by returning the mask of the desired object. \vspace{-1em}}
    \label{fig:sesame}
\end{figure*}

However, what if the ``keys with the purple heart'' do not exist in the scene? While recent methods for reasoning segmentation have shown remarkable performance on ``positive'' queries where the query object exists in the scene, most existing approaches for reasoning segmentation fail to account for this ``false premise'' scenario \cite{lai2023lisa}, and happily produce a hallucinated segmentation even when the objects associated with the query do not exist in the image \cite{wu2022towards, liu2023gres} (see \cref{fig:teaser}). It is desirable for robust reasoning segmentation systems to not only respond in the negative but to also propose corrected expressions when appropriate. 
Such robust systems should first be able to \textbf{``see''}, by detecting \textit{if} an object from the query is present in an image, then \textbf{``say''} something {\it about} the object itself if it's not there, suggesting alternatives to the user's query and optionally providing additional helpful information in the scene,
before finally being able to \textbf{``segment''} by showing {\it  where} in an image an object is grounded, if the user has not withdrawn their request.

Until now, ``false premise"-aware approaches have focused on the ``see'' and ``segment'' components, often using two-stage cascaded approaches, where an auxiliary classifier is used to ``see'' and a segmentation backbone is used to ``segment" \cite{wu2022towards, liu2023gres, wang2020referring}. These  pipeline-based approaches do not demonstrate reasoning ability when interpreting a reference and can not engage with users in task-directed dialogue; they cannot ``say'' anything about the query if it is incorrect,  unfortunately letting users continue with erroneous assumptions and queries without correction.

Addressing this unexplored area, we introduce a novel dataset and associated benchmarks, dubbed \datasetnameplain, \datasetnameplus, and \datasetnameg. These datasets, an expansion from RefCOCO(+/g) \cite{mao2016generation, yu2016modeling}, are augmented with context-aware false-premise queries via Large Language Models (LLMs), which are essential to train and evaluate a model's ability to ``see'', ``say'' and ``segment''. In this new task, we find that existing open-source LMM-based referring segmentation approaches often fail to ``see'' and ``say'' due to catastrophic forgetting during instruction fine-tuning.

To counter this issue, we develop two reasoning segmentation methods resilient to false premises. Our method includes a cascading models approach and an all-encompassing LMM, {\bf \ourmethodshortname{}} (\textbf{SE}e, \textbf{SA}y, seg\textbf{ME}nt), which is jointly trained with our novel dataset. By leveraging the reasoning and referencing nature capabilities of contemporary LMMs, these methods can not only ``see" and ``segment" but also ``say" what is necessary to reject or even correct a query. In summary, our contributions include the introduction of a novel benchmark dataset and:

\begin{itemize}
    \item \textbf{An LMM that can ``see'': } We analyze the ways in which existing approaches for reasoning segmentation fail to recognize false premise queries, and show that cascading models and joint data fine-tuning to produce relative accuracy improvements of up to \textbf{55.45\%} over a baseline's ability to detect false-premise queries.
    \item \textbf{An LMMs that can ``say'':} We further show our approach is novel in that it can give helpful feedback about the query, and demonstrate that such feedback is judged to be helpful up to \textbf{67\%} of the time. 
    \item \textbf{An LMM that can ``segment'':} Finally, we demonstrate the importance of false-premise robustness in improvements in segmentation quality, showing that robust false-premise training can result in relative cIoU improvements over baselines of up to \textbf{31.65\%}.
\end{itemize}

\section{Related Work}

\begin{figure*}[t]
\centerline{\includegraphics[width=1.0\linewidth]{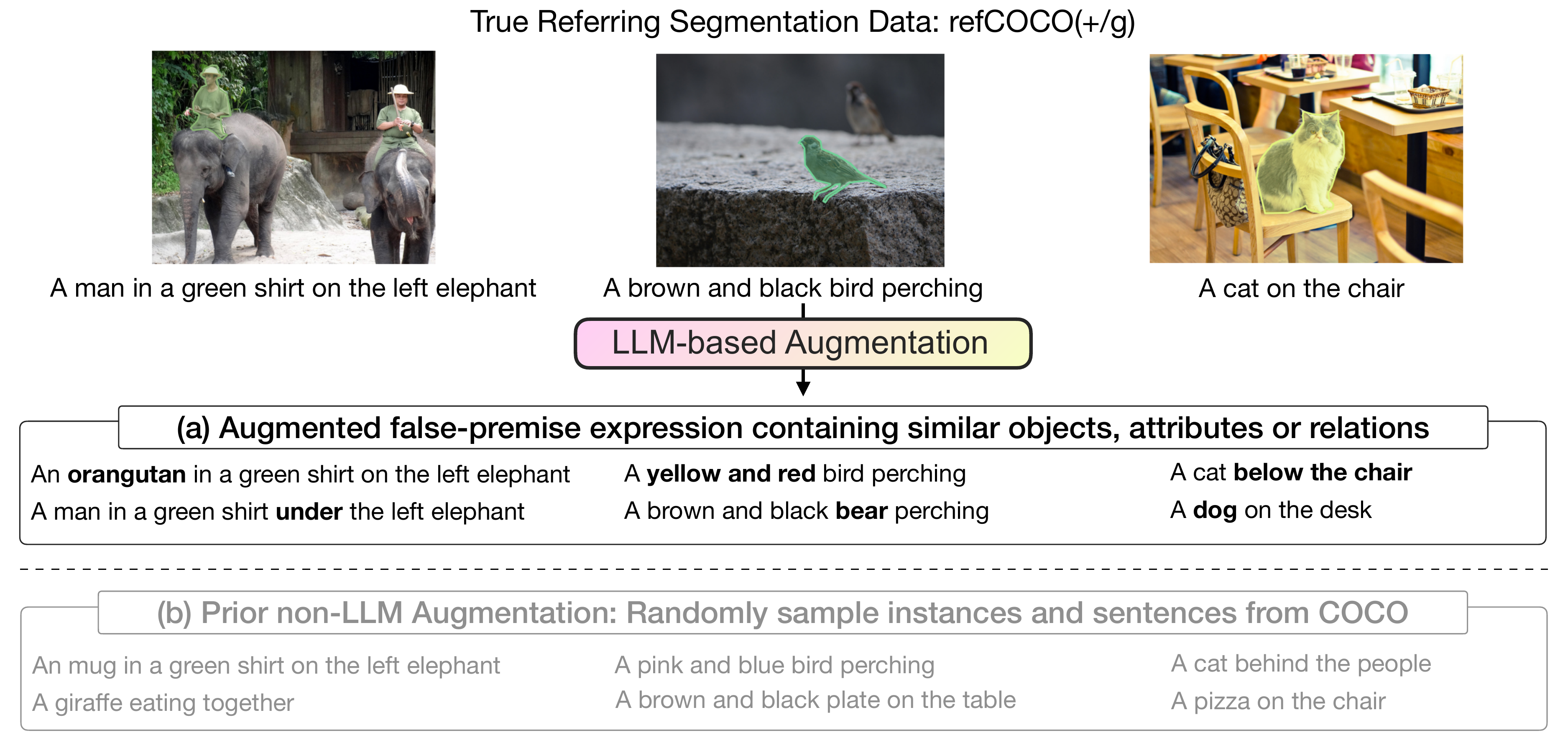}}
    \caption{\datasetname ~Dataset Creation. Using refCOCO for base images, we employ an LLM to create a false-premise referring segmentation dataset with similar objects, attributes, and relations. Such paired examples enable the the creation of specific correction ground truth that is more specific than baseline methods which simply sample positive and negative examples.  This data allows us to train an LMM that has robust reasoning reference capabilities. }
    \label{fig:dataset}
    \vspace{-1em}
\end{figure*}

Reasoning segmentation, a subset of referring segmentation, introduced by Lai \etal \cite{lai2023lisa}, focuses on complex reasoning tasks in addition to localized references. Reasoning segmentation exists in contrast to the more global tasks of semantic segmentation, which assigns class labels to every pixel in an image \cite{fcn, deconvnet, segnet, unet, deeplab, dilation, parsenet, pspnet, icnet, denseaspp, danet, ccnet, psanet, asymmetric_nonlocal, cheng2021per, lai2021semi, tian2022adaptive, tian2023learning}, instance segmentation, which detects pixels corresponding to instances of objects in a scene  \cite{he2017mask,zhang2021k,cheng2022masked}, and panoptic segmentation, which solve both instance and semantic segmentation problems simultaneously \cite{kirillov2019panoptic,xiong2019upsnet,cheng2020panoptic,li2021fully}. It is also more fine-grained than approaches for referring object grounding and reasoning \cite{peng2023kosmos, chen2023shikra, zhang2023gpt4roi, chen2023position, zhao2023bubogpt, zang2023contextual, you2023ferret, le2020dynamic, huang2022deconfounded, tang2023contrastive, ding2022vlt, yang2022lavt}, as these approaches operate on bounding boxes corresponding to objects, and do not seek to localize the pixels of the objects directly.

The current state-of-the-art for reasoning segmentation, LISA \cite{lai2023lisa}, uses a pre-trained large multimodal model  fine-tuned  to output segmentation tokens for each image. While LISA is capable of complex reasoning, it is trained in a manner which encourages producing segmentation/region outputs, even in the presence of a false-premise query. Similar to LISA, X-Decoder \cite{zou2023generalized} and SEEM \cite{zou2023segment} can both produce pixel-level segmentation and language tokens, however focus on multi-task performance, and struggle to perform complex reasoning segmentation tasks \cite{lai2023lisa}. 

While our proposed method is the first to explore false premises in the field of reasoning segmentation, understanding and detecting false premises has been studied in several other areas in computer vision \cite{davis2020unanswerable} including visual question answering \cite{ray2016question}, image/text matching \cite{feng2012automatic,xu2015show,ordonez2011im2text,karpathy2015deep,fang2015captions}, image-grounded conversation \cite{mostafazadeh2017image}, tool usage \cite{toor2019question} and hallucination detection \cite{rohrbach2018object}. Indeed, it has long been known in visual question answering that sometimes the image cannot entail any of the possible answers, and both datasets \cite{johnson2017clevr, suhr2017corpus, liu2019clevr, hudson2019gqa, gurari2018vizwiz, mahendru2017promise} and methods \cite{whitehead2022reliable, mashrur2023robust, mashrur2022semantic, li2020neural, prabhakar2018question, mahendru2017promise} have been developed which can evaluate and correct for false premises in the case of question answering.  Generally, methods for detecting false premises fall into a cascaded approach with two components, a ``detection'' model which is designed to determine if the question is answerable (``see''), and the standard ``answering'' model, which actually answers the question \cite{mashrur2023robust, mashrur2022semantic, li2020neural, wang2020referring, prabhakar2018question}.

Beyond question answering, several explicit measures have been designed which use pre-trained vision and language models to determine how closely text matches with a related image \cite{yi2020improving, cui2018learning, kane2020nubia, jiang2019tiger, hessel2021clipscore, clip, cho2023visual}, however while such measures can detect image/text relevance, they can neither ``segment'' nor ``say''. 

Closest to our work, Wang \etal \cite{wang2020referring} introduce a cascaded method for referring segmentation in the presence of false premises composed of an entity detection module, an expression parsing module (which parses objects using a textual scene graph), and a complex entity/relationship matching detection method based on WordNet distances. While this method is capable of understanding false premises and giving feedback in referring expressions, it cannot handle open-domain language commonly found in reasoning segmentation tasks and is restricted to closed-domain tasks with fixed vocabularies. Our work is the first approach which enables false premise detection and language feedback in open-domain reasoning segmentation tasks.

\begin{table*}[t]
\centering
\begin{tabular}{llrrrrr}
\toprule
Dataset & Split & Images & Objects & Sentences & \makecell{Positive Sentences} & \makecell{Negative Sentences} \\
\midrule
\multirow[t]{5}{*}{\textbf{\datasetname{}}} 
 & train & 16,992 & 42,278 & 234,445 & 120,191 & 114,254 \\
 & val & 1,500 & 3,805 & 20,962 & 10,758 & 10,204 \\
& testA & 750 & 1,975 & 11,205 & 5,726 & 5,479 \\
 & testB & 750 & 1,798 & 9,514 & 4,889 & 4,625 \\
\cmidrule{2-7}
 & \textbf{Total} & 19,992 & 49,856 & 276,126 & 141,564 & 134,562 \\
\cmidrule{1-7}
\multirow[t]{5}{*}{\textbf{\datasetnameplus{}}} 
 & train & 16,994 & 42,404 & 234,892 & 120,624 & 114,268 \\
 & val & 1,500 & 3,811 & 21,094 & 10,834 & 10,260 \\
& testA & 750 & 1,975 & 11,061 & 5,657 & 5,404 \\
 & testB & 750 & 1,810 & 9,883 & 5,095 & 4,788 \\
\cmidrule{2-7}
 & \textbf{Total} & 19,994 & 50,000 & 276,930 & 142,210 & 134,720 \\
\cmidrule{1-7}
\multirow[t]{4}{*}{\textbf{\datasetnameg{}}} 
 & train & 21,899 & 42,246 & 157,866 & 80,512 & 77,354 \\
 & val & 1,300 & 2,573 & 9,554 & 4,896 & 4,658 \\
 & test & 2,600 & 5,023 & 18,830 & 9,602 & 9,228 \\
\cmidrule{2-7}
 & \textbf{Total} & 25,799 & 49,822 & 186,250 & 95,010 & 91,240 \\
\bottomrule
\end{tabular}
\caption{Details for the \datasetname{} datasets, and train/val/test splits. Our dataset splits mirror those of RefCOCO (unc), RefCOCO+ (unc), and RefCOCOg (umd). To each original dataset, we have appended an approximately equal number of negative referring expressions, each coupled with a corresponding corrected sentence, thus creating an augmented dataset specifically for the False Premise Correction task.}

\label{tab:fprefcoco_splits}
\vspace{-1em}
\end{table*}

\section{A New Dataset and Benchmark for False Premise Correction}
\label{sec:taskanddataset}

Dialog-based models with the ability to segment and reason are traditionally trained on referring expression datasets which tend to only contain positive examples\textemdash examples that contain the object pertaining to the query language. Models trained under these conditions will always produce positive results, regardless of the truthfulness of the premise. Prior false-premise strategies to tackle this often integrate a classifier ahead of the segmentation module~\cite{liu2023gres,wu2022towards}, but this solution can be too restrictive, limiting the LMMs in engaging with diverse, open-domain conversational scenarios.

In response, \textit{we alter both the task and the data} to facilitate the ability for models to provide more human-like responses when presented with a question about a non-existent object. This new task, False Premise Correction, expects models to suggest an alternative referring expression that more closely matches an object in the image if prompted with a query that describes a missing object.

Although existing datasets such as R-RefCOCO \cite{wu2022towards} include queries referring to non-existent items in images, their method of generating negative expressions through naive random sampling often lacks context awareness. This limitation significantly reduces their effectiveness for false-premise correction tasks. Consider an image with a cat on a chair as in \cref{fig:dataset} (b): contextually valid false premises that could be logically corrected to ``a cat on the chair" might include phrases like ``a cat under the chair" or ``a dog on the chair." However, R-RefCOCO typically produces less suitable examples, such as ``a pizza on the chair” or ``a cat behind the people," which do not align with realistic model correction expectations. Furthermore, these datasets do not provide a direct link between each false premise query and its corresponding correct alternative, a critical aspect for effective training and evaluation in false premise correction.

To address these issues, we present \datasetname{}(+/g), a new benchmark dataset building upon the RefCOCO(+/g) referring segmentation datasets \cite{mao2016generation, yu2016modeling}. For each image, \datasetname{}(+/g) not only incorporates the original positive referring queries but also pairs them with a diverse range of contextually related false premise queries. To generate negative samples, we modify a single element (object, adjective, or relation) in the positive referring expressions by prompting the OpenAI GPT-3.5-turbo model \cite{brown2020language}. As depicted in \cref{fig:dataset} (a), our LLM-based augmentation strategy yields false premise queries that are more closely aligned with the context. After some basic data cleaning to ensure the responses are parseable, we end up with a nearly 1:1 positive/negative sample ratio and the same train/test/val splits as RefCOCO(+/g). Full statistics are provided in \cref{tab:fprefcoco_splits}.

The \datasetname{}(+/g) benchmark dataset enables the evaluation and training of Language and Multimodal Models (LMMs) in open-domain reasoning and segmentation tasks, focusing on three essential capabilities: ``See," ``Say," and ``Segment." In \cref{table:refer_seg} and \cref{table:abl}, the statistics showed significant limitations of the state-of-the-art model, LISA \cite{lai2023lisa}, particularly in its complete inability to reject non-existent items (``See") with 0\% recall on false premise query or to provide any appropriate corrections (``Say"). LISA predicts a segmentation for all false premise sentences, resulting in an approximately 30\% reduction in segmentation cIoU compared to the original dataset without any false premise queries. In response, we developed and trained an integrated LMM to achieve notable improvements across all three capabilities, which is detailed in \cref{sec:methods}.

\tabfpref{t}

\tabAbl{t}

\begin{figure*}[t]
    \centerline{\includegraphics[width=1.0\linewidth]{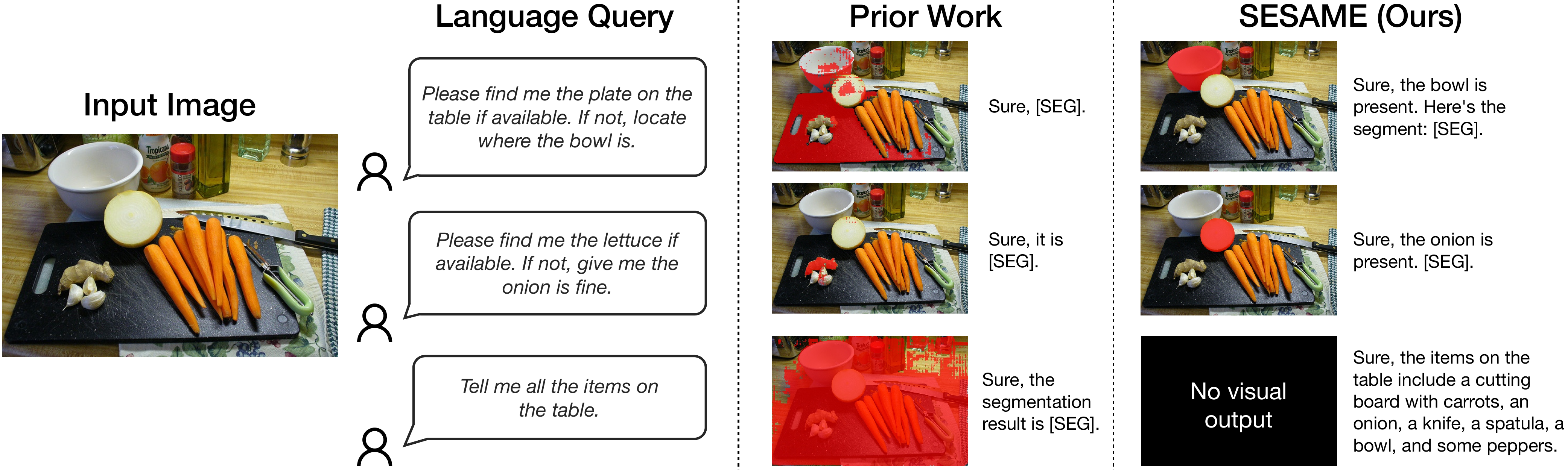}}
    \caption{In contrast to prior work (the output of the LISA \cite{lai2023lisa} is shown above), SESAME is able to handle more complicated conditional reasoning and instruction, and is able to not output a segment when it is not requested.}
    \vspace{-0.5em}
    \label{fig:multiround_conv}
\end{figure*}

\section{An LMM that can See, Say, and Segment}
\label{sec:methods}

To enable intelligent interaction systems that simultaneously possess the abilities to see, say, and segment, we first introduce a novel approach cascading various LMMs with distinct functionalities. We then present   \ourmethodshortname{}, a unified \textbf{SE}e, \textbf{SA}y, seg\textbf{ME}nt %
model with the aid of our curated dataset described above. %

Existing generic LMMs for VQA, such as GPT-4V \cite{gpt4v} or LLaVA \cite{liu2023visual}, are adept at identifying objects in images and suggesting alternatives when necessary. However, segmentation-specialized LMMs such as LISA, while capable of generating segmentation masks given diverse language prompts, struggle with queries about non-existent objects. In these cases, LISA often produces segmentation masks but fails to provide relevant feedback, typically offering generic responses like ``Sure, it is [SEG]." This behavior deviates from our desired outcome, indicating a need for more context-aware responses in advanced interaction systems. As shown in \cref{table:refer_seg}, there is a significant degradation in the performance of ``seeing'' and ``saying''. Intriguingly, the original LLaVA model which LISA is based on, prior to its fine-tuning for segmentation capabilities, did possess these abilities. This indicates a critical issue in the realm of LMMs – the challenge of catastrophic forgetting during the process of learning new skills.

To address this, we first propose a cascading approach for the False Premise Correction task. The first-stage LMM detects the presence or absence of objects in images and also engages in dialogue with users, providing clarifications or alternative suggestions when necessary. Once an object's presence is confirmed, the query can then be passed to the second LMM which specializes in the task of ``segmentation.''  This second-stage model is a segmentation-focused LMM \cite{lai2023lisa} that performs referring segmentation via prompts in the form of ``Please help me segment X in the image" and has high performance in both conventional semantic segmentation and complex reasoning segmentation tasks. This method coordinates between the two LMMs via prompt chaining with the first excelling in accurate object detection and contextual language response and the second in detailed image segmentation.

However, a single model with all three capabilities is desired, but as described above existing approaches ``forget'' how to ``see'' and ``say''. We address the catastrophic forgetting problem by utilizing a joint-training strategy. As in \cite{lai2023lisa} we instantiate \ourmethodshortname{} with LLaVa-v1.5 \cite{liu2023visual} for the ``see'' and ``say'' portions of the pipeline and Segment Anything \cite{kirillov2023segment} as the segmentation backbone. 

This training utilizes three distinct datasets: the train split (``train" from \cref{tab:fprefcoco_splits}) of the custom-designed \datasetname{}(+/g), the LLaVA VQA instruction finetuning dataset, and the train splits from R-RefCOCO(+/g). We call this the unified training set. The \datasetname{}(+/g) dataset comprises both positive and negative queries, with the incorrect ones being amended and always related to their original versions. In contrast, R-RefCOCO(+/g) includes randomly selected nonexistent COCO objects, which lack contextual relevance to the images; this dataset is employed to train the model to simply reject non-existent objects rather than offer corrections. 

We distributed \datasetname{}(+/g), LLaVA VQA, and R-RefCOCO(+/g) in a 7:2:1 ratio for each training cycle. For \datasetname{}(+/g) we specifically maintained a 9:1 ratio of true to false queries to ensure a balanced focus on the model's `say' and segmentation tasks. Training with R-RefCOCO(+/g) conditions the model to dismiss false premises without proposing alternatives. Relying solely on \datasetname{} could lead the model to generate speculative outputs, such as arbitrarily altering `left' to `right', without genuine image analysis. This issue is detailed in our ablation study (\cref{fig:abl}). Finally, the LLaVA VQA dataset, focused solely on visual question answering, is integrated to retain the model's competence in this field.

This simple yet effective approach enables the fine-tuned model to acquire new segmentation skills while preserving its innate ``see'' and ``say'' abilities. Unlike previous methods \cite{wu2022towards, liu2023gres} that used an auxiliary branch for binary responses to detect true/false premise queries, our approach seamlessly integrates the ``see'' and ``say'' abilities within the inherent capabilities of the LMM. This integration results in a more streamlined model capable of multitasking. The enhanced LMM demonstrates an improved ability to not only accurately segment objects in images while also engaging in intelligent dialogue, handling both existent and non-existent objects with equal finesse.

\section{Experimental Results}
\label{sec:experiments}

\begin{figure*}[t]
    \centering
    \vspace{-2em}
    \includegraphics[width=0.9\linewidth]{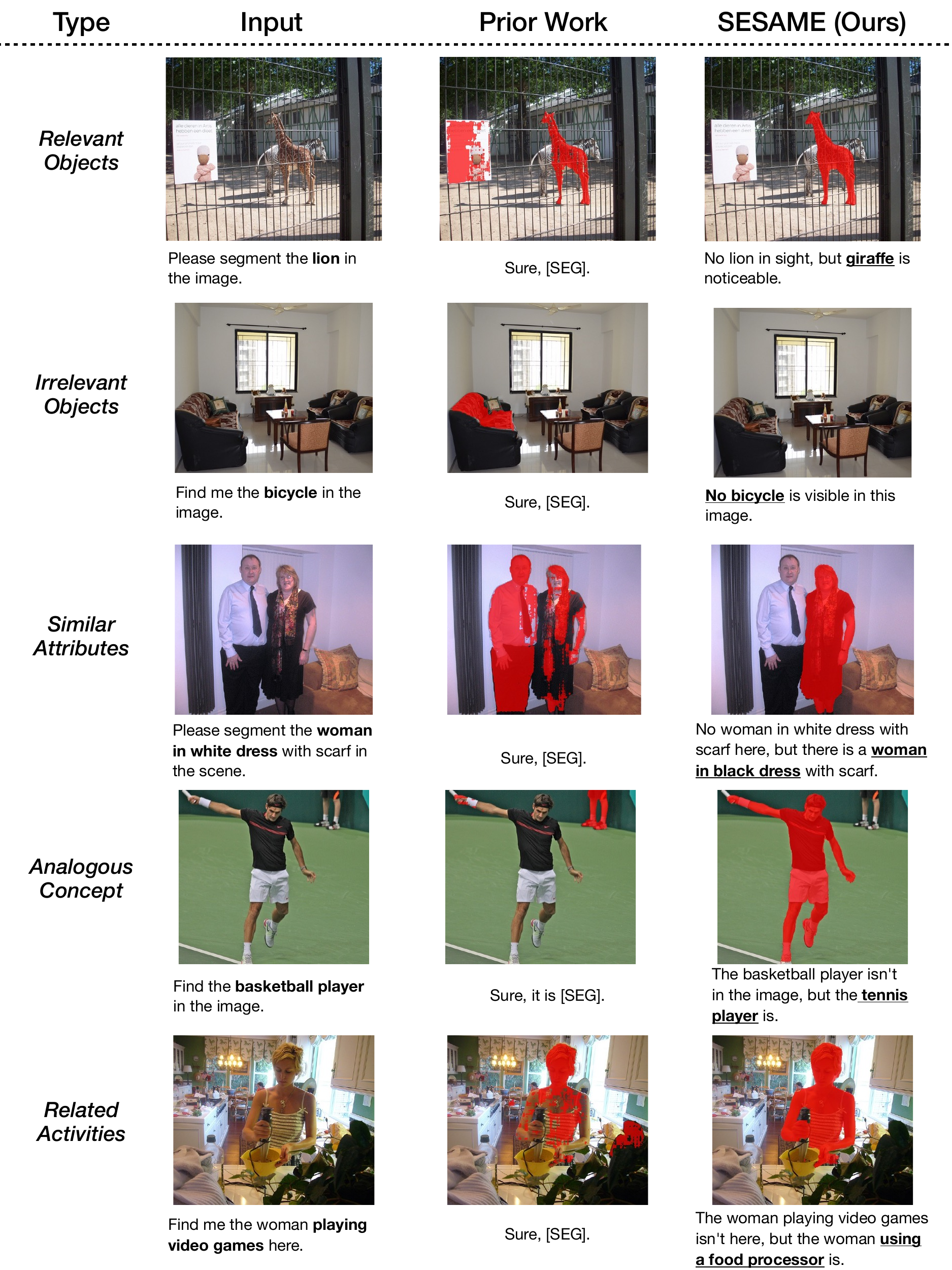}
    \caption{Not only is SESAME robust to false premises, and does not attempt to incorrectly predict a segmentation mask when an object or concept is not actually present in the image, but it is able to use commonsense reasoning to suggest relevant objects or concepts when a similar instance is present, and output the segmentation mask of that instance.}
    \label{fig:cherrypicking1}
\end{figure*}

\paragraph{Implementation Details.}
\ourmethodshortname{}'s LMM backbone, LLaVA-v1.5-7B, and segmentation decoder, SAM, are fine-tuned for 10 epochs via LoRA \cite{hu2021lora} using the unified training dataset described in \cref{sec:methods}. Our loss function combines dice and binary cross-entropy losses for segmentation, along with cross-entropy loss for sentence prediction tasks. We employed the AdamW optimizer with a learning rate of 3e-4, setting the batch size to 10 and the gradient accumulation steps to 5. The total training time was approximately 24 hours on a single DGX A100 80GB GPU. To ensure fair comparisons, following \cite{lai2023lisa}, we carefully excluded images from the training set that were also present in the test or validation sets. This step was crucial to avoid data contamination, especially as we merged FP-RefCOCO, FP-RefCOCO+, and FP-RefCOCOg into a unified training set, each having unique data splits. We will make our code available for future research and applications.

\subsection{Results}
In our experiments, we assessed the ``See'', ``Say'', and ``Segment'' capabilities of \ourmethodshortname{}, our cascading method (combining the off-the-shelf LLaVA-v1.5-7B and LISA-7B), and the baseline model LISA-7B. Results are reported on the test sets of \datasetname{}(+/g) in \cref{table:refer_seg}.

\paragraph{Detection.} \vspace{-0.6em} Models' ability to ``See'' is evaluated using binary classification accuracy, a metric that determines whether models can accurately discern the presence or absence of an object referenced in an image. In this evaluation, \ourmethodshortname{} achieves an accuracy of 79.84\%, outperforming both the cascading method (75.59\%) and LISA-7B (51.4\%). LISA-7B, expectedly, underperforms due to its constant erroneous assumption that the prompted object exists in the scene. The superior detection accuracy of \ourmethodshortname{} when compared to the cascading method stems from our enhanced fine-tuning approach. This approach integrates a more extensive and balanced collection of both positive and negative referring expression training data, unlike the standard VQA dataset used in the conventional LLaVA-v1.5 fine-tuning process. %

\paragraph{Description.} \vspace{-0.3cm} We evaluate a model's ability to ``Say'' through a modified CLAIR metric \cite{chan2023clair}. CLAIR uses an LLM (OpenAI's GPT-4 \cite{OpenAI2023GPT4TR}) to provide scores for candidate captions compared against a reference caption set, outputting a similarity score in $[0, 1]$. We use the model's corrected referring expression as the candidate and compare it against the set of positive sample referring expressions belonging to the same referent in the image. We modified CLAIR to return the score of the best match to the reference set. When there are multiple objects with referring expression annotations in the same image, we only score the suggested correction against expressions for the same object instance.

The cascading method, employing LLaVA for the ``say" function, demonstrates superior performance, surpassing the finetuned LISA which appears to lose all the capability. This finding underscores the problem in LISA's fine-tuning process. In contrast, \ourmethodshortname{} achieves much higher CLAIR scores ($0.63$, $0.61$, $0.67$) than the cascading method ($0.35$, $0.42$, $0.55$) on all three datasets. We theorize this is because \ourmethodshortname{} is fine-tuned on our customized false premise data with ground truth false premise corrections that encourage the model to learn to suggest the same referent rather than other irrelevant objects in the image.

\paragraph{Segmentation.} \vspace{-0.4cm}The segmentation performance is measured using the cIoU metric, following established protocols in previous conventional and false-premise referring segmentation studies \cite{wu2022towards,liu2023gres, lai2023lisa}. This metric assesses the cumulative intersection over cumulative union. As expected, as \ourmethodshortname{} has the best ``See" ability, it consistently skips the creation of segmentation masks for false premise queries, thereby achieving the highest cIoU scores. Conversely, the baseline LISA model, which produced segmentation masks for all false premise queries, recorded the lowest cIoU scores.

\begin{figure}[t]
    \centering
    \includegraphics[width=\linewidth]{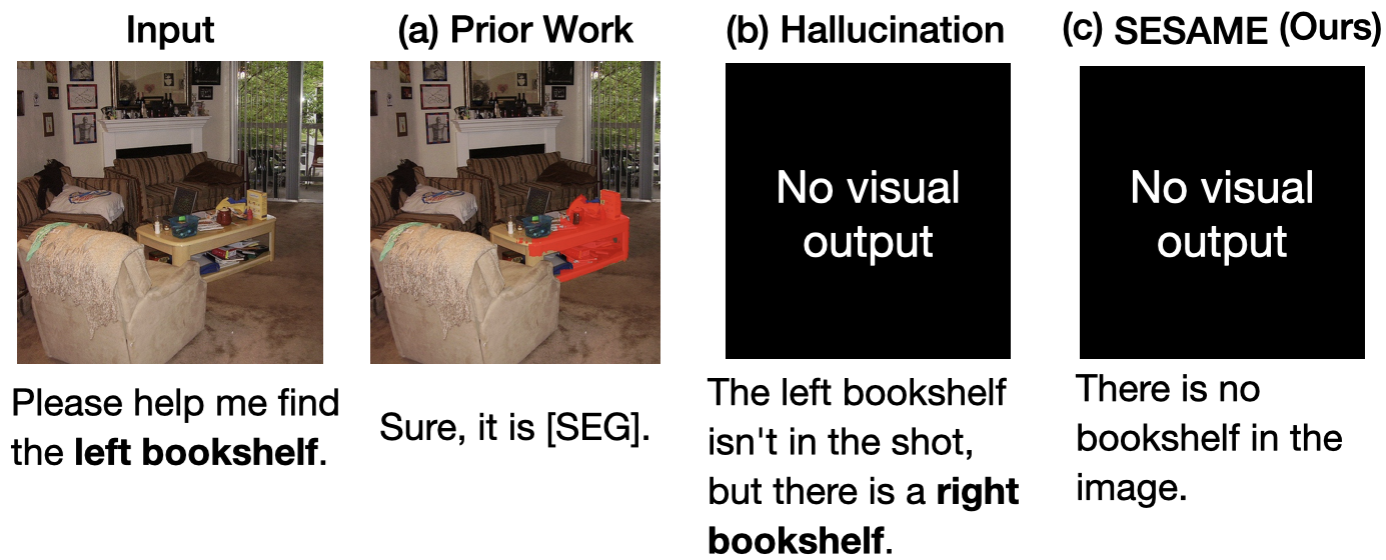}
    \caption{Ablation Studies: (a) Prior work hallucinates a segmentation even though there is no bookcase. (b) If we force the model to correct all the false premise queries, it correctly detects that there is no left bookcase in the image, but still hallucinates a ``right bookcase'' in the text response, likely because relational expressions are often reversed (e.g., ``left'' to ``right'') when modified to form negative samples. (c) \ourmethodshortname{} addresses the hallucination problem by adding R-RefCOCO(+/g) into the unified train set, and allowing the model to simply respond that the requested object was not found rather than requiring it to provide an alternative expression. 
    \label{fig:abl}
}
\vspace{-1em}
\end{figure}

\subsection{Discussions}

\paragraph{Proportion of False-premise Queries.}  An essential part of our analysis involved varying the proportion of false premise queries in our test set, as detailed in \cref{table:abl}. A 0\% false premise (FP) scenario is equivalent to the RefCOCOg evaluation, while a 50\% FP mirrors the \datasetname{g} dataset. These findings underscore that despite fine-tuning, models, including \ourmethodshortname{}, still have considerable potential for improvement, specifically in detecting false premises (FP) with the highest Recall in false premise query being only 67.89\%. This capability is particularly crucial as increasing FP proportions directly impacts the performance of downstream segmentation cIoU score.

\paragraph{Handling Complex Instructions.} \vspace{-0.5cm} A particularly notable example in \cref{fig:multiround_conv} showcases \ourmethodshortname{}'s ability to process and respond to complex user prompts. This includes segmenting an alternative object based on a conditional query and engaging in VQA only without generating a segmentation mask. In contrast, previous models like LISA lacked these capabilities, severely limiting their human-like interaction potential. This finding also suggests that \ourmethodshortname{} could be extended to multi-round interactions, where a user might request an intelligent agent to first summarize a scene and then focus on segmenting specific objects of interest.

\paragraph{Significance of False Premise Rejection.} \vspace{-0.5cm} We also investigated the impact of integrating the R-RefCOCO(+/g) dataset, specifically designed for false premise rejection, into our training process. Excluding these data often led models to rely on superficial word modifications in their responses instead of genuinely interpreting the image context. This reliance aggravated the issue of hallucination and resulted in lower scores in the ``Say" capability. A striking illustration of this phenomenon is presented in \cref{fig:abl}.

\vspace{-2em}

\paragraph{Performance in Established Referring Segmentation Benchmarks.}  Finally, we assessed \ourmethodshortname{} in traditional referring segmentation benchmarks with only positive queries. The results in \cref{tab:segmentations} demonstrated the comparison between our method and several existing approaches. While our model is adept at handling false premises and enhancing dialogue interaction, it does not compromise the segmentation abilities. This indicates that our joint training approach, which fine-tunes LMMs with a tailored dataset, successfully achieving great segmentation  capabilities while maintaining robust performance in LMM's basic ability to see and say.

\tabNatref{t}

\section{Conclusion}

In this study, we tackle the overlooked issue within the realm of LMMs: false premise segmentation queries. We not only highlight this challenge in existing LMM methodologies but also introduce a pioneering task known as False Premise Correction, necessitating capabilities to ``See," ``Say," and ``Segment." Alongside this new task, we present \datasetname(+/g), a specially designed dataset for evaluating LMMs on these essential skills. To address this challenge, we employ innovative cascading and joint training techniques. Our integrated LMM, \ourmethodshortname{}, demonstrates substantial improvement in detecting the presence of objects ("see"), advising users about non-existent objects or modifying queries accordingly (``say"), and precisely segmenting objects that are actually present in the image (``segment"). This research fills a critical gap in LMM capabilities and sets a strong foundation for future explorations into improving LMM interactions in diverse and real-world applications.

\section*{Acknowledgements}
This work was supported in part by funding from the DoD and from the BAIR Commons program.
{
    \small
    \bibliographystyle{ieeenat_fullname}
    \bibliography{main}
}

\clearpage
\appendix

\section{Extension to Reasoning Segmentation Tasks}

Our main paper highlights the effectiveness of \ourmethodshortname{} in answering queries, with or without false-premises, for open-language segmentation tasks. The method operates by ``seeing'' whether the referred object is present in an image, ``saying'' what the correct grounding is, and ``segmenting'' the image using the appropriate input prompt. While our initial focus is on false-premise referring segmentation tasks, our method also encompasses reasoning segmentation tasks.

Reasoning segmentation tasks \cite{lai2023lisa} represent a more complex challenge compared to traditional semantic or referring segmentation tasks. These tasks necessitate advanced reasoning and world knowledge---models need to understand complex queries with intricate expressions or longer sentences. In such scenarios, models are tasked with not only identifying objects in an image but also comprehending and reasoning about the broader context and the relationships depicted within the scene.

\vspace{1em}

\begin{figure*}[t!]
    \centerline{\includegraphics[width=1.0\linewidth]{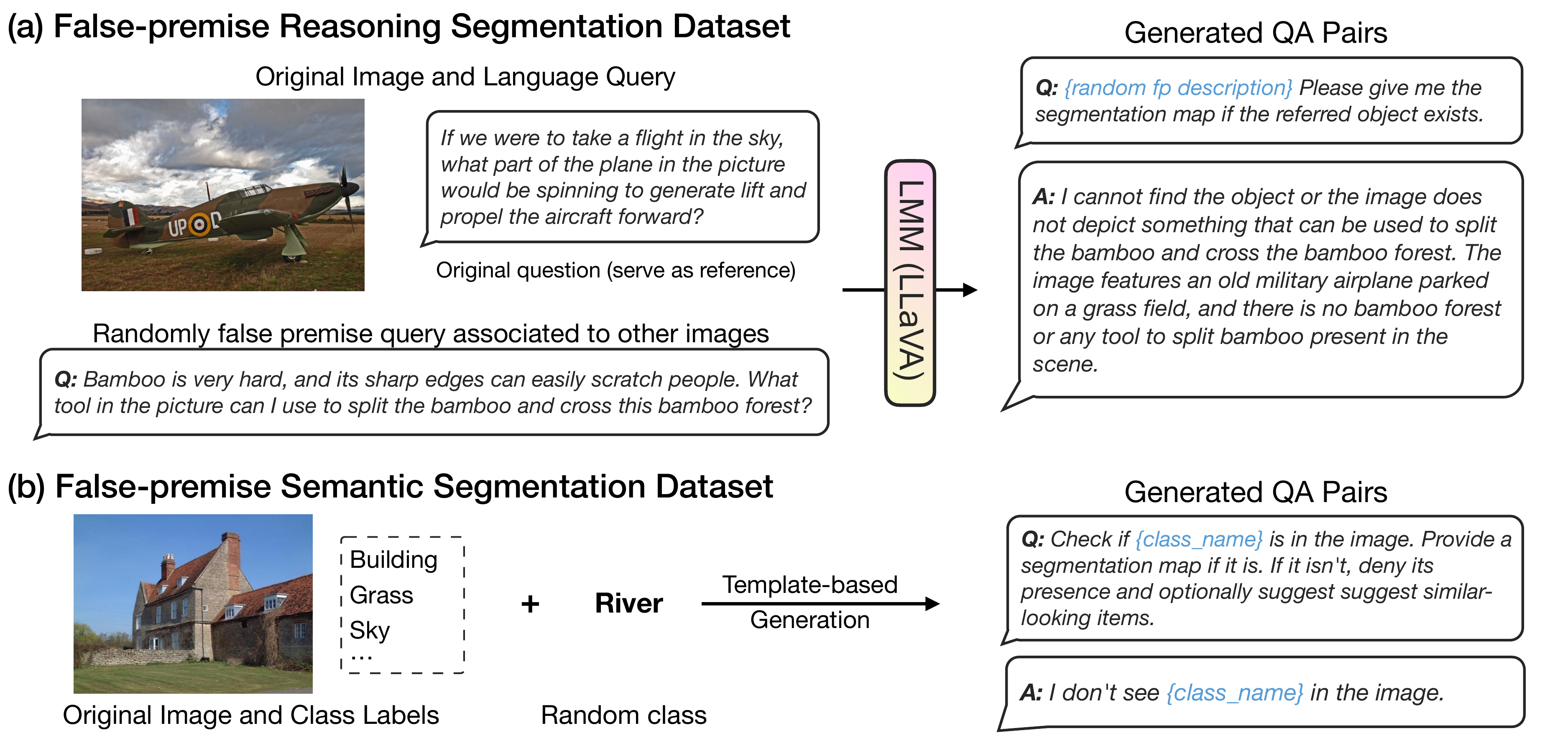}}
    \caption{(a) We employ LLaVA to generate false-premise question-answer pairs for the reasoning segmentation dataset. (b) We use a template-based method to create false-premise semantic segmentation data from well-known datasets, including ADE20K, COCO-stuff, and LVIS-PACO part segmentation. Leveraging these datasets, along with the unified dataset which encompasses our curated FP-RefCOCO(+/g) datasets as detailed in \cref{sec:methods}, \ourmethodshortname{} effectively demonstrates its proficiency in handling false-premise queries within the realm of reasoning segmentation tasks as shown in \cref{table:reasoning_seg}.}
    \label{fig:supp_data}
\end{figure*}

\noindent \textbf{Setup.} In line with the method outlined in \cref{sec:methods}, we initially developed a specialized dataset for false-premise reasoning segmentation, which includes both training and validation components. This dataset, with an equal number of false-premise queries and the original true queries, was derived from the original one proposed by \cite{lai2023lisa}. As shown in \cref{fig:supp_data} (a), for each image within this dataset, we randomly selected language queries that correspond to other images, along with one original question and its corresponding image. These elements were then used as inputs for LLaVA \cite{liu2023visual} which was tasked with generating plausible question-and-answer pairs.

For the training of \ourmethodshortname{}, we incorporated several datasets. These include ADE20K \cite{zhou2017scene}, COCO-stuff \cite{caesar2018coco}, and LVIS-PACO part segmentation \cite{ramanathan2023paco} for semantic segmentation in addition to the reasoning segmentation dataset and the unified dataset as described in \cref{sec:methods} of the main paper. Specifically, for the semantic segmentation datasets, we utilized a template-based method to create false-premise query-answer pairs, as depicted in \cref{fig:supp_data} (b). The training split for the reasoning segmentation dataset followed the previously detailed procedure.

The architecture of the \ourmethodshortname{} model is the same as described in \cref{sec:methods}. We allocated the training dataset in a ratio of 10:3:1, spanning the semantic segmentation dataset, the unified dataset (as detailed in \cref{sec:methods}), and our augmented reasoning segmentation dataset. We deliberately employed a 9 times higher sampling rate for true premise queries compared to false-premise ones to ensure effective training for segmentation tasks. The training adheres to the same hyper-parameters and procedures as detailed in \cref{sec:experiments}.

\tabSupp{h!}

\vspace{1em}

\noindent \textbf{Quantitative Evaluation.} When evaluating LMMs with our augmented official validation set of the reasoning segmentation dataset, we primarily focus on assessing the ``See'' and ``Segment'' components of our method. Given the inherent challenges in evaluating the ``Say'' component using the CLAIR score \cite{chan2023clair} — which typically relies on explicit expressions similar to captions (e.g., ``a trash can") rather than the implicit references characteristic of reasoning segmentation tasks (e.g., ``the place we throw the garbage") — we decided not to include this component in our quantitative analysis. However, to illustrate this aspect of our model's capabilities, qualitative examples are provided in the subsequent section.

In evaluating the ``seeing'' ability of our model, we reported the recall for both true and false premise queries concerning object existence in the image. Segmentation performance was measured using the cIoU metric, consistent with the metrics utilized in Table 2 and Table 3 of our main paper. The results of this evaluation are presented in Table \cref{table:reasoning_seg}, where we compare our \ourmethodshortname{} method with both the LISA baseline and our proposed chained method.

The data reveals that while the chained method outperforms the LISA baseline, our \ourmethodshortname{} method shows even more significant advancements. Specifically, it achieved an impressive 12 times increase in object detection (See) scores for false premise queries with a slight decrease in performance for true premise queries. \ourmethodshortname{} also demonstrated a notable improvement in segmentation performance, exceeding the chained method and the LISA baseline by over 7\% in terms of cIoU. These findings highlight the critical importance of incorporating false-premise queries into model training and illustrate the wide-ranging effectiveness and adaptability of our method across various segmentation subtasks.

\vspace{1em}

\begin{figure*}[t]
    \centerline{\includegraphics[width=0.9\linewidth]{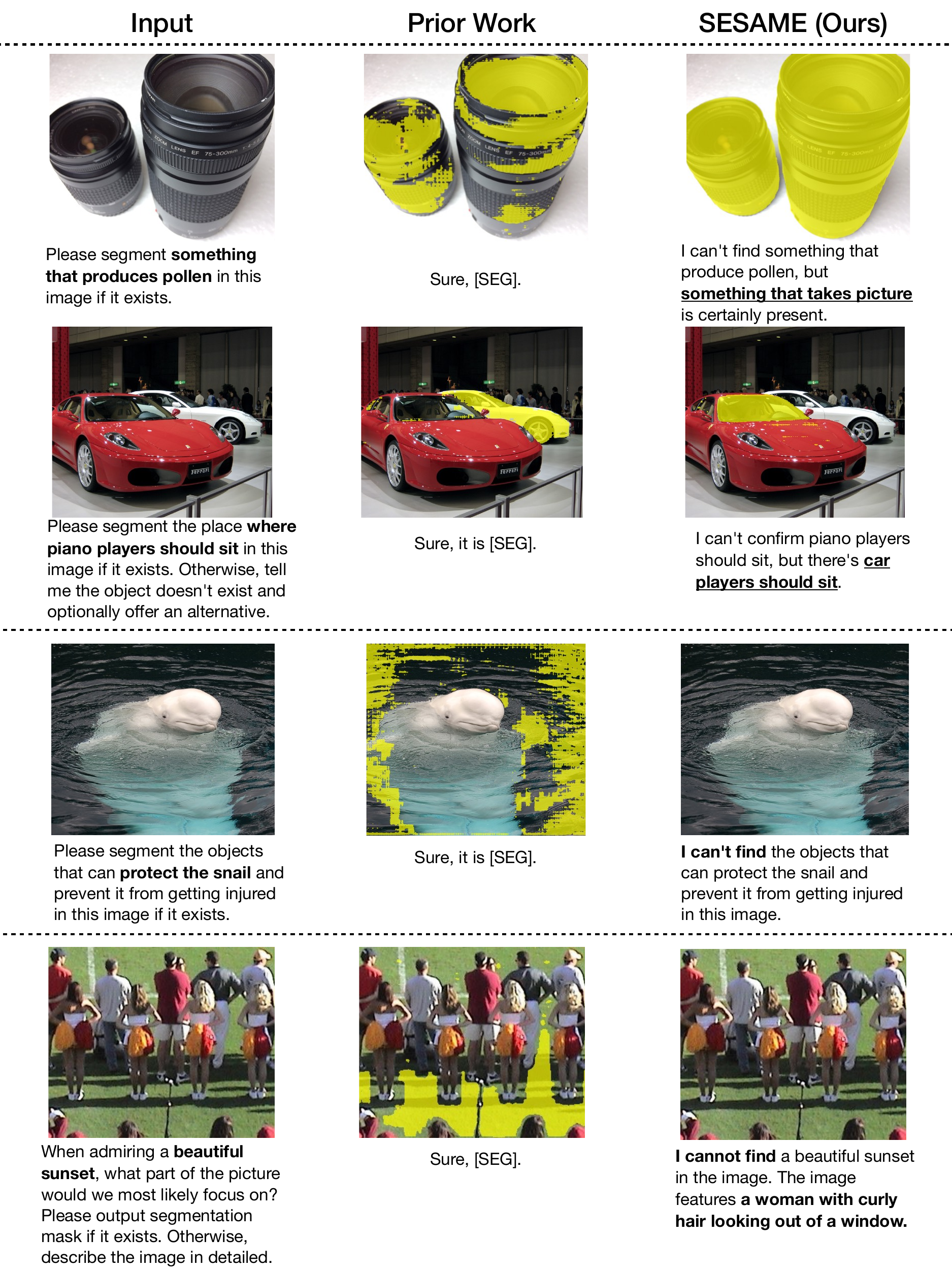}}
    \caption{In contrast to prior work (the output of the LISA \cite{lai2023lisa} is shown above), \ourmethodshortname{} typically succeeds in rectifying complex input queries and refrains from generating a segment when unnecessary.}
    \label{fig:supp_say}
\end{figure*}

\noindent \textbf{Qualitative Results.} As illustrated in \cref{fig:supp_say}, we showcase a selection of question-answer pairs generated by \ourmethodshortname{} on the reasoning segmentation dataset. These examples demonstrate the model's proficiency in the ``Say" component, particularly in addressing complex queries. 

In the first and second rows, \ourmethodshortname{} successfully identifies that the referred objects do not exist in the image and provides appropriate suggestions in response. This ability to accurately deny the existence of objects and offer relevant alternatives showcases a significant advancement over prior methods. 

In the third example, when presented with a query that is entirely irrelevant, \ourmethodshortname{} demonstrates its ability to appropriately deny the query without offering unnecessary suggestions. This highlights the model's nuanced understanding and response generation capability. In contrast, the LISA baseline method, when presented with false premise queries, tends to generate segmentation maps without the capacity to offer alternative responses in all cases. 

However, as indicated in the fourth row, there is room for improvement. While \ourmethodshortname{} exhibits a strong ability to deny non-existent objects, it occasionally generates hallucinated corrected results. This means that the model can occasionally generate a ``corrected'' premise that is still false. Addressing this issue will be a focus of our future work, aiming to enhance the model's accuracy and reliability in generating corrected premises.

\begin{figure*}[t]
    \centering
    \includegraphics[width=0.9\linewidth]{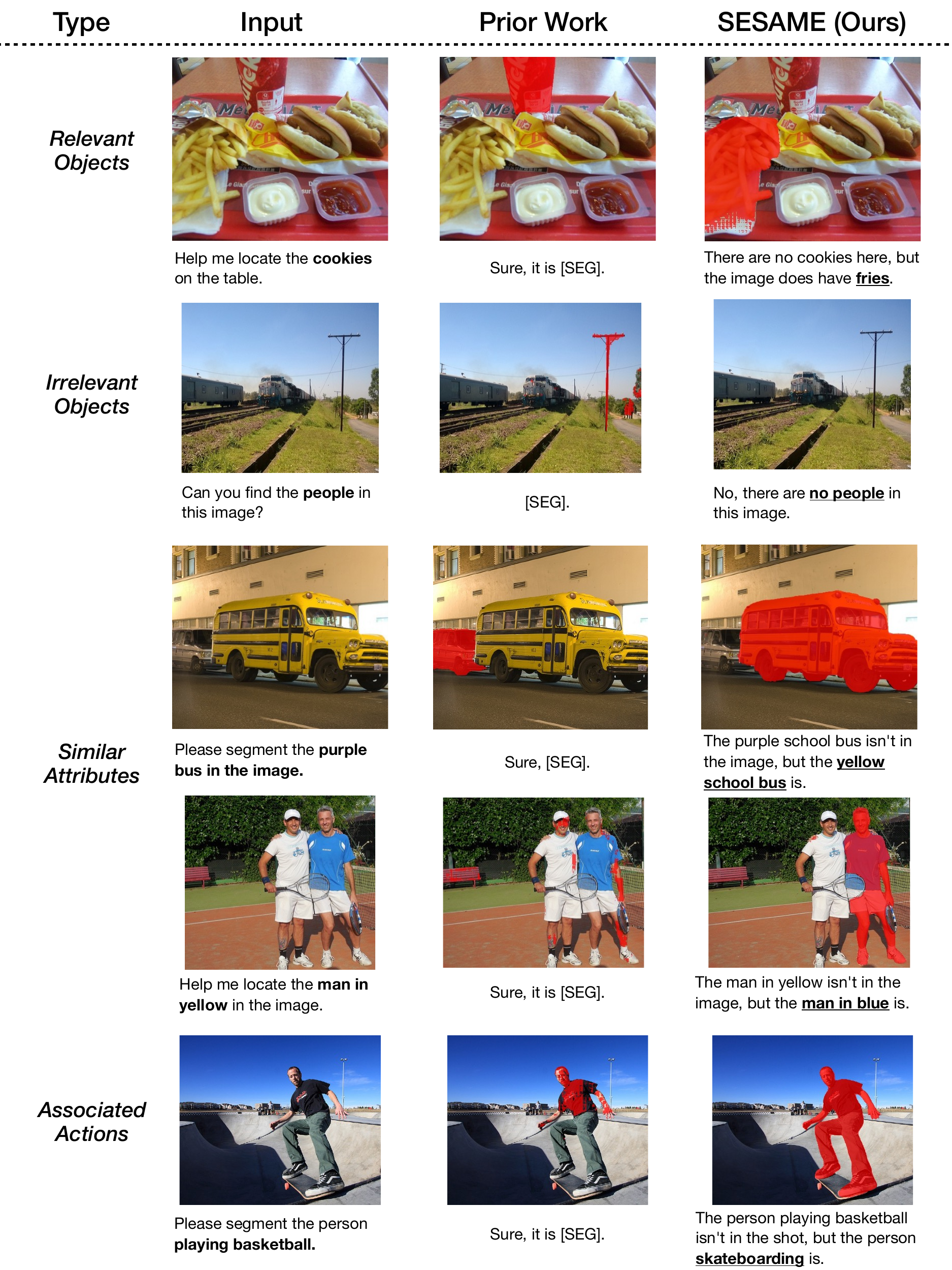}
    \caption{\ourmethodshortname{} not only robustly identifies and refrains from predicting segmentation masks for non-existent objects or concepts in an image but also intelligently suggests and outputs segmentation masks for similar, relevant instances through commonsense reasoning.}
    \label{fig:supp_cherrypick}
\end{figure*}

\section{More Qualitative Results}

In \cref{fig:supp_cherrypick}, we present additional visual comparisons between \ourmethodshortname{} and previous methods in the context of referring segmentation tasks. A notable strength of \ourmethodshortname{} lies in its capacity to reject and, where appropriate, correct queries founded on erroneous assumptions. This capability is particularly evident in instances where the actual object bears a certain relevance to the falsely presumed item in the query, encompassing similarities in objects, attributes, or actions.

\section{Our Prompts to the LLM and LMM}

Our method is anchored by two specialized prompts. The first, as elaborated in \cref{tab:llava_prompt}, underlies our cascading method. Specifically, we input the prompt into LLaVA-v1.5 to obtain the ``see and say'' results. A key aspect of this process involves identifying non-existent predictions: if the output begins with ``No, there is no...,'' we classify these as non-existent predictions and collect their corresponding sentences to assess the ``say'' score. This prompt, designed to elicit explanatory responses through chain-of-thought prompting techniques \cite{wei2022chain}, has shown significantly greater efficacy compared to simpler user prompts like ``Please help me segment [the referring expression] in the image?''

The second prompt used in our dataset development is outlined in  \cref{tab:our_full_prompt}. This prompt includes carefully crafted instructions and utilizes a chain-of-thought approach \cite{wei2022chain} in its in-context examples. This design is important to minimize the generation of false premise sentences that either deviate significantly from the input object or reference an existing one.

\begin{table}[h!]
\setlength\tabcolsep{0pt}
\centering
\begin{tabular*}{\linewidth}{@{\extracolsep{\fill}} l }\toprule
\begin{lstlisting}[style=myverbatim]
Analyze the image and verify if there are any referred objects in the image. Yes or no with explanations. Here's an example. When asked "is there any green car behind the man in the image", you can answer the question in ways like:
                
1. If the object exists, confirm it:
	- "Yes, the green car hehind the man is present in the image."

2. If not, deny the existence of the object and optionally provide alternative suggestions:
	- "No, there is no green car in the image. Did you mean the red car in front of the man?"

Now, my question is: "Is there [the referring expression] in the image?" I value a precise and detailed analysis. Please inspect the image thoroughly and respond according to the guidelines provided above.
\end{lstlisting} \\\bottomrule
\end{tabular*}
\caption{Our full prompt for the LLaVA-v1.5 model to obtain the result of see and say.}
\label{tab:llava_prompt}
\end{table}

\begin{table*}[h!]
\setlength\tabcolsep{0pt}
\centering
\begin{tabular*}{\linewidth}{@{\extracolsep{\fill}} l }\toprule
\begin{lstlisting}[style=myverbatim]
## Your Role: Prank Expert

## Objective
Turn real object descriptions of an image into fictional but relevant ones by altering specific elements.

## Guidelines
- Change only one word in each sentence.
- Use unique word replacements in each sentence.
- Focus on modifying the main subject, its attributes, actions, or relationships to another relevant counterpart.
- Ensure altered descriptions do not coincide with any real objects in the original description.
- Be cautious when changing adjectives related to position (e.g., near/far, left/right) and size (e.g., small/large) to avoid ambiguity and inadvertent overlap with existing items.

## Example 1:
Original: ["The red ball to the left of the blue toy.", "The man in a white shirt standing next to a woman with an umbrella.", "The smallest dog in the group, near the tree."]
Altered: ["The yellow ball to the left of the blue toy.", "The woman in a blue shirt standing next to a woman with an umbrella.", "The smallest cat in the group, near the tree."]
Reasoning: Changes focus on the main subject (ball color, person's gender, animal type) while ensuring uniqueness and avoiding overlap with real objects.

## Example 2:
Original: ["a man getting ready to cut a cake", "guy in green with a knife in the right hand picture", "woman pointing at ice cream", "a woman in a blue shirt with floral print", "the man standing up and pointing"]
Altered: ["a kangaroo getting ready to cut a cake", "guy in purple with a knife in the right hand picture", "woman pointing at a pizza", "a woman in a blue shirt with stripes", "the man standing up and stretching"]
Reasoning: Each alteration (animal for person, color change, relevant object swap, pattern change, action change) ensures a distinct and fictional transformation while maintaining the sentence structure.

## Example 3:
Original: ["the vase on the right", "a woman wearing a blue shirt", "a woman in a grey blue sweatshirt painting a figure onto a vase", "man on right", "man in red shirt", "a large brown urn being decorated by a woman", "a large urn on a green table and the lady is drawing on it"]
Altered: ["the vase in the back", "a baby wearing a blue shirt", "a woman in a grey blue sweatshirt jogging on the sidewalk", "man on the top", "man in orange shirt", "a large school bus being decorated by a woman", "a large urn on a couch and the lady is drawing on it"]
Reasoning: The changes made here (position, subject identity, action, location) are carefully chosen to create fictional scenarios without referring to other existing objects in the original description.

Your Turn Now! Adhere to the guidelines and answer the question!

Original: [A list of referring sentences of a single COCO image]
Altered: 
\end{lstlisting} \\\bottomrule
\end{tabular*}
\caption{Our full prompt to the GPT4 for augmenting false-premise referring expression. LLM starts completion from ``\texttt{Altered:}''.}
\label{tab:our_full_prompt}
\end{table*}

\end{document}